\documentclass[runningheads]{llncs}
\usepackage{preamble2}

\begin{document}
\title{Breaking the Symmetries of Indistinguishable Objects}
\author{{\"O}zg{\"u}r Akg{\"u}n\inst{1}\orcidID{0000-0001-9519-938X} \and
Mun See Chang\inst{1}\orcidID{0000-0003-2428-6130} \and
Ian P. Gent \inst{1}\orcidID{0000-0002-5604-7006}\and
Christopher Jefferson\inst{1,2}\orcidID{0000-0003-2979-5989}}
\authorrunning{Akg{\"u}n et al.}
\institute{School of Computer Science, University of St Andrews, UK  \and
School of Science and Engineering, University of Dundee, UK \\
\email{\{ozgur.akgun, msc2, ian.gent, caj21\}@st-andrews.ac.uk}}
\maketitle

\begin{abstract}
Indistinguishable objects often occur when modelling problems in constraint programming, as well as in other related paradigms.   
They occur when objects can be viewed as being drawn from a set of unlabelled objects, and the only operation allowed on them is equality testing.  
For example, the golfers in the social golfer problem are indistinguishable. If we do label the golfers, then any relabelling of the golfers in one solution gives another valid solution.  Therefore, we can regard the symmetric group of size $n$ as acting on a set of $n$ indistinguishable objects.   In this paper, we show how we can break the symmetries resulting from indistinguishable objects.   We show how symmetries on indistinguishable objects can be defined properly in complex types, for example in a matrix indexed by  indistinguishable objects.  We then show how the resulting symmetries can be broken correctly.  
In \essence, a high-level modelling language, indistinguishable objects are encapsulated in `unnamed types'.  
We provide an implementation of complete symmetry breaking for unnamed types in \essence. 
%In this paper, we show how we can break the symmetries resulting from unnamed types. 
%However, the method can be prohibitively expensive to do, as we shall see. So we also study methods for breaking the symmetry only partially but much more efficiently.
%These methods should generalise to any other problems where there are indistinguishable objects. 
    
\keywords{Symmetries \and  Modelling  \and  Constraints programming \and Automated model transformations}
 \end{abstract}

\section{Introduction}

Symmetries have long been understood to be both widely occurring and a source of inefficiency for solving technologies. 
As a result, this has been an exceptionally well-studied topic in constraint programming~\cite{GENT2006329}, Boolean satisfiability~\cite{SAThandbook}, and mixed-integer programming~\cite{margot2009symmetry}.
% and related areas~(see, for example, 
% \cite{GENT2006329,SAThandbook,margot2009symmetry,Achterberg2013}).
A particularly important case of symmetries is where the problem has indistinguishable objects.
These are objects which, when interchanged, give us essentially the same situation. 
For example, two machines of the same model are equivalent in a factory scheduling problem, and any valid schedule will give an equivalent schedule when two such machines are interchanged. 
Further complications are introduced when we have multiple sets of indistinguishable objects, and we are not allowed to interchange objects of the different sets. 

When modelling problems with symmetries, due to the limited choices of representations, one tends to \emph{introduce} symmetries that are not in the original problem. These symmetries must then be broken manually, e.g. by adding symmetry breaking constraints. High-level modelling languages such as \essence allow one to specify the problems in a more abstract way, and symmetries introduced by modelling can then be handled automatically by \conjure, an automatic model rewriting tool. 
In \essence, `unnamed types' were introduced to capture the notion of indistinguishable objects~\cite{bakewell2003towards}. Compound types can be constructed in terms of these unnamed types, for example, we can have sets of tuples of indistinguishable objects. 
%Due to the complexity of breaking these symmetries, the automatic generation of symmetry breaking constraints for unnamed types has been largely neglected. 
However, while unnamed types were present in the first version of \essence, previously \conjure ignored the symmetry of unnamed types and simply transformed them into integers. 
As we shall see in this paper, handling these symmetries is significantly more difficult than other symmetries already managed by \conjure.

To handle the symmetries of indistinguishable objects, we extend \essence to support permutations of types. These permutations allow us to capture, in \essence, the symmetries of indistinguishable objects. This gives a general and extensible framework for breaking the symmetries of objects constructed from indistinguishable objects. 
By showing how we implement permutations, we show how other technologies that want to deal with the symmetries of indistinguishable objects can adopt a similar approach. 
As the types of variables including unnamed types can be arbitrary nested, and the list of objects we want to handle might be extended in the future, we give the semantics of \essence types recursively, in terms of a much smaller set of mathematical objects. 
We use this semantics to define how symmetries of indistinguishable objects induce symmetries of objects constructed from them. 
We also show how a well-defined total ordering on a compound type can be built up in terms of the ordering of its constituent types. 

These ingredients let us generalise the lex-leader symmetry breaking method to compound types. 
We will illustrate our technique in the context of \conjure, but we will also discuss how other modelling languages can take the same approach to remove symmetries due to indistinguishable objects. 
%Then the lex-leader constraints for a simple type can be lifted to form the lex-leader constraints of a compound type. 

Often we do not want to break all of the symmetries in a model since that encoding would be detrimental to performance. 
For this reason, we also explore weaker, partial forms of symmetry breaking, offering a modelling choice between fast and complete symmetry breaking. 
We show, using well-known constraint models containing unnamed types, how our symmetry breaking encodings can be applied to them using this abstraction. 
As an example, we show that the commonly used `double-lex' method \cite{rowColSym} naturally arises from our methods.

In~\Cref{sec:background}, we give a brief overview of symmetry breaking in constraint programming and the \essence language, and then define the symmetries of unnamed types in \Cref{sec: unnamed types general section}. 
We shall define the symmetries induced by unnamed types and see how we can break them, using a newly defined total order of the values of any type. 
We shall also see how several relaxations give us some incomplete but faster symmetry breaking constraints.  
In~\Cref{sec: implementation}, we describe the method is implemented in \conjure and in \Cref{sec: discussion} give some case studies.

\subsubsection{Motivating Example: Social Golfers Problem}
\label{section: motivating example}

The social golfers problem \cite{csplib:prob010} asks for a schedule for $p$ people playing golf over $w$ weeks in $g$ groups per week. To attain maximum socialisation, the constraint is that no two different golfers play in the same group as each other in two different weeks.  
Notice that the problem only cares whether or not two golfers are the same person or different people.  That is, the golfers are all \textit{indistinguishable}.  Similarly, the groups are also \textit{indistinguishable}, and so are the weeks.  For example, the constraints take no account of weeks being consecutive or otherwise: two weeks are either identical or distinct.  This means that given any solution to the problem, we can obtain another solution through any permutation of the golfers. Similarly, we can  also permute the groups  and the weeks freely.  Furthermore, we can apply any permutation of the golfers, groups and weeks independently to get another solution.  With so many symmetries in hand, manually breaking them may require some modelling expertise, whether the aim is to break all symmetries completely or to do some form of partial symmetry breaking.  
The approach we take in this paper allows this to be done automatically.  We will do this in the context of \conjure, working on the \textit{unnamed types} allowed by \essence, in this case for {golfers}, {weeks} and {groups}.  Our approach is general and could be applied in other modelling languages.

\section{Background}
\label{sec:background}

A \emph{constraint satisfaction problem} (CSP) $\mathcal{P}$ with $n$ variables is a triple $(V,D,C)$, where $V = \{V_1 , V_2 , \ldots, V_n \}$ is the set of \emph{variables}, 
$D$ consists of sets $\Dom(V_i)$, called the \emph{domain} of $V_i$, for each $1 \leq i \leq n$, 
and $C = \{ C_1, C_2,  \ldots, C_k \} $ is the set of \emph{constraints}, where each $C_i$ is a subset of the Cartesian product $\bigtimes_{1 \leq i \leq n} \Dom(V_i)$.  %do i need the airities
An \emph{assignment} of the variables in $V$ is a $n$-tuple $(a_1, a_2, \ldots,  a_n) $, where each $a_i \in \Dom(V_i)$. 
An assignment is a \emph{solution} to $\mathcal{P}$ if it is in the intersection $ \bigcap_{1 \leq i \leq k} C_i$. 
The \emph{solution set} to $\mathcal{P}$ is the set of all solutions to $\mathcal{P}$.

A \emph{permutation} of a set $\Omega$ is a bijection from $\Omega$ to itself. We typically denote permutations using the cycle notation. That is, a permutation $\sigma: =(a_{11},a_{12}, \ldots, a_{1k_1}) $$(a_{21},a_{22}, \ldots, a_{2k_2})$ $ \ldots (a_{r1},a_{r2}, \ldots, a_{1k_r})$ means that, for all~$i$, we have $a_{ij} \mapsto a_{i(j+1)}$ for $j<k_i$ and $a_{ik_i} \mapsto a_{i1}$.  
The composition of two permutations $\sigma_1$ and $\sigma_2$ is denoted by $\sigma_1 \sigma_2$, the inverse of a permutation $\sigma$ is denoted by $\sigma^{-1}$. 
For $\omega \in \Omega$ and a permutation $\sigma$ over $\Omega$, we denote the image of $\omega$ under $\sigma$ by $\omega^\sigma$. 
A \emph{permutation group} $G$ over a set $\Omega$ is a set of permutations over $\Omega$ that is closed under compositions and inverses. 
Such $G$ necessarily contains the identity $1_G$, the permutation that fixes all points in $\Omega$. 
The set of all permutations over $\Omega$ is called the \emph{symmetric group} of $\Omega$, denoted by $\Sym(\Omega)$. 

A \emph{group action} of a group $G$ on a set $X$ is a map $\phi: G \times X \rightarrow X$ such that $\phi(1_G,x) = x$ for all $x \in X$ and $\phi(gh, x) = \phi(h, \phi(g, x))$. 
When there is a group action, we may write $x^g$ for $\phi(g,x)$. 
This aligns with the notation of permutation application because $\Sym(\Omega
)$ naturally acts on $\Omega$. 
When there is such a group action, we say that $G$ \emph{acts on} $X$ and that $G$ is a \emph{symmetry group} of $X$.

When we have a symmetry group acting on the domains of a decision variables $V$, we have an equivalence relation on the domain set $D$ (and hence the set of all solutions), where two assignments $(a_1, a_2, \ldots, a_n)$ and $(a'_1, a'_2, \ldots, a'_n)$ are equivalent if there exists $g \in G$ such that $a_i^g=a'_i$ for all $1 \leq i \leq n$. 
A \emph{symmetry breaking constraint} is a constraint where, when added to the constraint set, removes or reduces symmetric values from consideration. 
A \emph{sound} symmetry breaking is one where at least one solution from each equivalence class is outputted, while a \emph{complete} symmetry breaking only outputs exactly one per equivalence class.
Many symmetry breaking strategies use \emph{lexicographic ordering}. For a total ordering $\leq$ on set $A$, the \emph{lexicographical ordering $\leq_{lex}$ over $\leq$} is the total ordering on the tuples/lists/words/matrices over $A$ such that $(t_1, t_2, \ldots, t_k) \leq_{lex} (t'_1, t'_2, \ldots, t'_k)$ if and only if either $t_i = t'_i$ for all $i$, or there is an $i$ such that $t_i \leq t'_i$ and $t_j = t'_j$ for all $j < i$.

%\subsubsection{Related work}
Symmetry breaking for constraint programming is very well studied, and there are many papers on this topic (see \cite{GENT2006329} for an overview). 
Two closely related concepts are interchangeability of values \cite{freuder1991interchangeableVals} and intensional permutability of variables \cite{roy1998interchangeableVars}, but they only concern value and variable symmetries respectively.  
Our work differs from these as we take a type-directed view of indistinguishable objects, which  means that types with interchangeable values can be used to build higher-level types. 
Depending on how we build these higher-level types, the symmetries can be variable or value symmetries, or indeed both or neither, depending on where the unnamed types occur. 
This suggests that a new, more general, method of reasoning with symmetries is needed. 
%\conjure does automatic removal of symmetries introduced by reformulations, and here we are doing this too. relieving modeller to add their own sym break constriants. Unnamed types are in there but symmetries are not broken. 
Symmetries are introduced by \conjure when abstract types in \essence are refined to lower-level types in \essencePrime (see \Cref{sec: essense overview}). Currently all but the symmetries arising from unnamed types are automatically broken by \conjure (see \cite{akgun2022conjure} for more details). 
The symmetries in the constraint modelling language MiniZinc is also extensively studied (see, for example, \cite{baxter2016symmetry,mears2011proving}), but our work here differs in that we consider higher-level abstract types.

\subsection{\essence as a Modelling Language}
\label{sec: essense overview}

\essence and \essencePrime are both constraint specification languages. 
The domains of decision variables in an \essence or \essencePrime problem specification are defined by adding attributes and/or bounds to built-in types.
For example, in \essence, we can have a variable of domain \texttt{set (size 3) of matrix indexed by int(1..5) of bool}. 
In this paper, matrices indexed by $[I_1, I_2, \ldots, I_k]$ refers to $k$-dimensional matrices where the values are accessed by values of $I_1 \times I_2 \times \ldots \times I_k$, so an entry of such a matrix m is $m[i_1, i_2, \ldots, i_k]$, where each $i_j$ is in $I_j$.

%The type of such a variable is a set of matrix of Booleans, and putting the restrictions on sizes gives us its domain. 
Types in \essence and \essencePrime are divided into two kinds: atomic and compound. 
The atomic types of \essencePrime are Booleans and integers, while \essence further supports enumerated types and unnamed types.
Compound types, such as matrices, are defined using atomic types and can be arbitrarily nested. 
\essencePrime only support the matrix compound domain, while \essence also supports tuples, records and variants. 
\essence further supports abstract decision variables of set, multiset, sequence, function, relation and partition. Non-abstract domains are also called concrete domains. 
For more details, see~\cite{akgun2022conjure} or the documentation at {\url{https://conjure.readthedocs.io}}.

\begin{remark}\label{def: types summary}\label{def: semantics of types} \label{remark: values of types as sets and tuples}
For a type $T$, we denote its set of all possible values of $T$ by $\Val(T)$, which is defined in terms of matrices, multisets and tuples to avoid large case splits over all \essence type:
the values of \texttt{bool}, \texttt{int} and \texttt{enum} are what one would expect; 
the values of a \texttt{tuple}, \texttt{record}, \texttt{variant} and \texttt{sequence} can be naturally defined as tuples; the values of a \texttt{matrix} are matrices,
the values of a \texttt{set}, \texttt{mset}, \texttt{partition} can be naturally defined as (nested) multisets; for types $\tau_i$, 
the values of a \texttt{function $\tau_1 \rightarrow \tau_2$} are  subsets of $\Val(\tau_1) \times \Val(\tau_2)$ such that there are no two elements with the same value in its first position; 
the values of a \texttt{relation $(\tau_1 * \tau_2 * \ldots * \tau_k)$} are subsets of $\Val(\tau_1) \times \Val(\tau_2) \times  \ldots \times \Val(\tau_k)$. 
Note that the representations presented here give abstract meaning to the types, carefully selected to simplify the theory, but may not adhere to the representation of the underlying implementations. 
\end{remark}

There are two advantages for defining $\Val(T)$ in terms of matrices, \linebreak multisets and tuples. Firstly, this allows us to avoid large case splits over types, e.g. when defining the symmetries of compound objects built from unnamed types. Furthermore, this makes our method more general and applicable across other platforms. As long as we define the values of a type in a similar way, either for a new type in \essence or a type in other systems, the method described in this paper still applies. Note also that we can also have all types to be defined as multisets and tuples only. This is because a matrix $m$ indexed by $[\tau_1, \tau_2, 
 \ldots, \tau_k]$ of $\tau$ can be represented in terms of multisets and tuples as $\{ (i_1, i_2, \ldots, i_k, m[i_1, i_2, \ldots, i_k])  \mid i_j \in \tau_j \}$.

\conjure transforms a problem specification (a \emph{model}) in the \essence language into a problem specification in the \essencePrime language, through a series of rewrites or transformations. 
%A \emph{refinement} is the transformation of a problem specification (a \emph{model}) $M_E$ in the \essence language into  either another problem in \essence, or a problem specification $M_{E'}$ in the \essencePrime language. 
Concrete domains are those that can be represented directly in \essencePrime, possibly by separating into their components. 
%Here, enumerated types are transformed into integers, while tuples, records, and variants are separated into multiple variables of their components. 
These transformations are straightforward and do not introduce any symmetries (see~\cite{akgun2022conjure} for more information). 
The abstract types are removed in a series of rewrites called \emph{refinements}. 
%Each refinement consists of two steps: the representation selection, which dictates how an abstract domain is to be represented using a concrete domain; and the expression refinement, which concerns the rewriting of the constraints regarding the abstract domains according to the representation selection. 
For each abstract domain, 
%(\texttt{set}, \texttt{mset}, \texttt{sequence}, \texttt{function}, \texttt{relation}, \texttt{partition}), 
there is a choice for how to translate to a concrete domain. Such a choice is what we call a \emph{representation} of the abstract domain. 
The constraints regarding the abstract domains according to the selected representation. 
In some cases, an abstract type is represented as a concrete type that satisfies certain constraints (e.g. sets as lists with all different elements). Such constraints are called \emph{structural constraints}. 
representations used in \conjure are well studied 
We will not detail representations used in \conjure here, but  will describe the ones which we use in examples. 
We direct interested readers to~\cite[Table~5]{akgun2022conjure} for a summary of the representations in \conjure. 

\emph{Modelling symmetries} are introduced if we translate abstract decision variables to a variable with bigger domain, which may result in the increase of solution number.  
In \conjure, modelling symmetries without unnamed types are \textit{always} broken completely. 
%When this happens, we have any underlying \textit{bijective} map between the refined values (that satisfies certain structural constraints) and the unrefined values.
The complete symmetry breaking constraint for the refinement of each abstract type in \essence is well studied. 
For more details on refining abstract variables, please refer to \cite{akgun2022conjure}.

\section{Unnamed Types and How to Break Them}
\label{sec: unnamed types general section}

To model indistinguishable objects, we use the concept from \essence of `unnamed types'.  While \essence provides unnamed types as a primitive and we implement our techniques in \conjure, our work applies generally.  For any modelling situation to which unnamed types apply, the techniques we propose can be used whether or not the modelling language used provides unnamed types.  The advantage of having unnamed types as a primitive, as in \essence, is simply that no additional work is necessary to recognise the existence of indistinguishable objects.
In this section we show the value in modelling with unnamed types, discuss the symmetries inherent in unnamed types and how to break them.

\subsection{Modelling with Unnamed Types}
\label{sec: example models}

We shall briefly show some examples of how one would use unnamed types in modelling, to illustrate their usefulness for high-level modelling and the difficulties in breaking their symmetries. 
%Note that the actual models used in \Cref{sec: results} uses more abstract \essence types. 
Note that we can model these problems using more abstract \essence types, but we choose to avoid these abstract types in this paper in hope to better illustrate the potential use of unnamed types.

%\subsubsection{Social golfer problem}
Recall the \textbf{social golfer problem} from \Cref{section: motivating example}.
%The {social golfer problem }\cite{csplib:prob010} asks for a schedule for $p$ golfers playing in $g$ groups per week, over $w$ weeks such that, to attain maximum socialisation, no pairs of golfers play in the same group twice. 
A model may have a \texttt{matrix indexed by [int(1..w), int(1..p)] of int(1..g)} as a decision \linebreak variable, together with constraints for maximal socialisation and to make sure that the group sizes are as expected, which we shall omit here.  
As we have seen, the golfers, weeks and groups can be permuted while still giving us valid schedules.
% However, the golfers can be permuted while still giving us valid schedules. This corresponds to permuting the columns of the decision variable. 
% Similarly, the weeks can be permuted as well, giving us row symmetries. 
% Not only that, the group numbers do not matter -- permuting, say, group number 1 and group number 2 globally in the schedule still gives us an equivalent schedule. This means that the entries in the decision matrix can be permuted as well, as long as we apply the permutation consistently across the whole matrix. 
% With so many symmetries in hand, manually breaking them may require some modelling expertise. 
% %For example, the modeller would need to know about double-lex to break the row and column symmetries efficiently~\cite{rowColSym}. The symmetry of the entries adds another layer of complexity. 
Unnamed types can be used to express a set of objects whose labels are not important. That is, permuting the labels will give us an equivalent assignment (and hence solution). The labels for golfers, weeks and groups, encoded as integers here, do not matter -- permutations of them, when done consistently, will give us equivalent solutions. 
So we can define \texttt{golfers}, \texttt{weeks} and \texttt{groups} as unnamed types of size $p, w$ and $g$ respectively, and have the decision variable be \texttt{matrix indexed by [weeks, golfers] of groups}. Then \conjure will handle the symmetries from unnamed types automatically. 
%There is a choice to break all symmetries, which may be slow, but the incomplete version aligns with expertly crafted symmetry breaking constraints (see \Cref{sec: results}).

%In fact, in \essence, we can encode the decision variable as a $w$-set of partition of golfers. This way, the symmetries from weeks and the groups are handled natively as symmetries of sets and partitions, and we do not require the structural constraint to ensure the correct group sizes. This way, we can use unnamed types to deal with the symmetries of golfers only. See ?? for the final model. 

%\todo{want to say unnamed types has no order, so when getting pairs of golfers for the max socialisation constraint, we can't do $ g1 < g2 $ like in the conjure paper, figure 1.}

%\subsubsection{Template design problem}
The \textbf{template design problem}~\cite{csplib:prob002} arises in a printing factory that is asked to print $c_1, c_2, \ldots, c_k$ copies of designs $d_1,  d_2, \ldots, d_k$ respectively. 
Designs are printed on large sheets of paper and each piece of paper can hold at most $s$ designs. 
A template is defined by the designs to be printed on a paper (at most $s$ of them, can be repeated, order does not matter). 
Given a number $n$, we want to find $n$ templates $t_1,t_2, \ldots, t_n$, and the number of copies for each of them to satisfy the printing order, while minimising the total number of printing. 
One might model this problem with two decision variables.
Firstly, to encode the number of copies needed for each template, we use a \texttt{matrix indexed by [int(1..n)] of int} -- we call this matrix $M_1$.
Secondly, to encode the number of copies of each design in each template, we use a \texttt{matrix indexed by [int(1..n), int(1..k)] of int(1..s)} (called $M_2$). 
%, recognising that we cannot have more than $s$ copies of a design in any template. 
As before, the labels of the templates, currently integers from $1$ to $n$, do not matter -- permuting the labels gives equivalent solutions. Since the templates are used as indexes in two decision variables, any symmetry handling of one must be consistent with the other. For example, enforcing that $M_1$ must be sorted and at the same time enforcing that rows of $M_2$ are sorted may give us a wrong result. 
An alternative would be to replace \texttt{int(1..n)} in the indices for both $M_1$ and $M_2$ with an unnamed type of size $n$. Using unnamed types, \conjure shall automatically and consistently break the symmetries. 
%As in the social golfers example, our model in \Cref{sec: results} uses abstract \essence types such as functions instead of matrices. 
%We choose to present a model with matrices here to distinguish between the uses of abstract \essence types and unnamed types

%\subsubsection{Set-theoretic Yang-Baxter equation} 
The \textbf{set-theoretic Yang-Baxter problem} asks for a special class of solutions to the infamous Yang-Baxter equations, which gives insights to various subfields of algebra and combinatorics (see \cite{akgun2022YB} for references). A special class of the set-theoretic solutions of the Yang-Baxter equation can be modelled as a mapping $\varphi: X \times X \rightarrow X$ which satisfies certain constraints (see \cite{akgun2022YB} for details), for a set $X$. 
As is common in mathematics, the elements of such a set $X$ are unlabelled and interchangeable. For example, in this case, if $X$ is realised as a concrete set of $\{x_1, x_2, \ldots, x_n\}$, then any permutation of the elements of $X$ in the definition $\varphi$ gives another mapping that is essentially the same (or equivalent). 
As, again, is common in mathematics, we want to count the number of solutions up to equivalences, which means that we want to remove the symmetries due to the interchangeability of the elements of $X$. 
Such a map $\varphi$ can therefore be modelled as a \texttt{matrix indexed by [T,T] of T}, where $T$ is an unnamed type. 
In this problem, the same unnamed type is used both as indices (twice) and elements of a matrix, so swapping two values in $T$ requires swapping two rows and two columns, and also all occurrences of two values for all variables. 
In this case, the symmetries of this matrix is neither a variable variable or a value symmetry (see \cite{GENT2006329}), the two most well-studied families of symmetry in constraint programming. Expressing a symmetry breaking constraints for this symmetry requires significant expertise in constraint modelling, which limits the ability of many constraint users to deal with the symmetries of their problems.

\subsection{Symmetries of Unnamed Types}
\label{sec: unnamed types}

As we have seen above, when modelling, we often want to express that two items are equivalent or indistinguishable from each other. 
%An example of this is the golfers in the Social Golfer Problem,  where one is asked to schedule games for a number of golfers. 
In \essence, we model these using \emph{unnamed types}, which are sets of known size with implicit symmetries: values of an unnamed type are unlabelled and hence interchangeable.
The values of unnamed types are not ordered and the only operations allowed on unnamed types are equality and inequality. Unnamed types are atomic so they can be used to construct compound (concrete or abstract) domains in many ways, including the members of a set, the image or result of a function, or the indices of a matrix.
%This means that a model with unnamed types is implicitly telling the solver that there are symmetries. It is now up to \conjure to handle/break these symmetries automatically. 

In order to define symmetries inhabited by variables constructed from unnamed types, we need to first define the set of possible values of this variable. This proves to be difficult as, as soon as we enumerate its values, we have put a label on its elements and hence introduced symmetries. 
We therefore define unnamed types as an enumerated type that comes with a symmetry group: 

\begin{definition}\label{def: unnnamed type}
An \emph{unnamed type $T$ of size $n$} is an \essence type with value set 
$\Val(T) = \{1_T ,2_T , \ldots, n_T \}$, together with a symmetry group $\Sym(\Val(T))$ consisting of all permutations of $\Val(T)$. 
\end{definition}

For example, in the social golfer problem, we can represent three possible golfers as an unnamed type $G$ of size $3$. Then $\Val(G)$  is $ \{ 1_G,2_G,3_G \}$. 
Note that a value of an unnamed type is unique to its type. For example, $1_T$ can only be a value of an unnamed type $T$, and not at the same time be a value of another unnamed type $U$. This means that the values of distinct unnamed types $T$ and $U$ are always distinct. 
To simplify notations, we write $\Sym(T)$ for $\Sym(\Val(T))$.

As we are dealing with indistinguishable objects, the symmetry group in \Cref{def: unnnamed type} is a symmetric group. However, our method of breaking these symmetries completely never uses the property of symmetric groups. So the method is generalisable to the case where we have an atomic type $T$ with any permutation group $G \leq \Sym(\Val(T))$ on its values as its symmetry group. 

\subsubsection{Induced Symmetries on Compound Types}

When outputting solutions to a problem with unnamed types, we only want to retain the solutions up to symmetries. 
In the social golfer example, a solution is equivalent to another assignment where we have swapped two of the golfers, and so we should only output one of them. 
However, a decision variable can be an arbitrarily nested construction of various \essence types, which may also involve multiple unnamed types. 
So next we define how the symmetries of unnamed types induce symmetries of the compound variables constructed from unnamed types: 

\begin{definition}
\label{def: induced action}
Let $T$ be an unnamed type of size $n$ and $X$ be a compound type. 
Then $\Sym(T)$ acts on $\Val(X)$, via the action
\begin{enumerate}
    \item if $x$ is a value of type $T$, then $x^g$ is the image under the action on $\Val(T)$; 
    \item if $x$ is atomic and $x$ is not of type $T$, then $x^g=x$;
    \item if $x$ is a matrix indexed by $[I_1, I_2,  \ldots, I_k]$ of $E$ then for each $v \in \Val(X)$, the image $x^g$ is a matrix where its $i$-th element $y[i] $ is $(x[i^{(g^{-1})}])^{g}$, where $i^{(g^{-1})}$ denotes the preimage of $i$ under $g$; 
    \item if $x$ is a multiset $\{ v_1, v_2, \ldots, v_k\}$, then $x^g = \{v_1^g, v_2^g, 
 \ldots, v_k^g\}$; 
    \item if $x$ is a tuple $ (v_1,v_2, \ldots, v_k)$, then $x^g = (v_1^g, v_2^g, \ldots, v_k^g)$.
\end{enumerate}
%The action on other types can be determined using \Cref{remark: values of types as sets and tuples}. 
\end{definition}

Recall from \Cref{def: types summary} that possible values of a non-atomic variable can be constructed from only matrices, multisets and tuples. So \Cref{def: induced action} \emph{does} in fact define the image of all possible types in \essence, by deducing from \Cref{def: induced action}, and considering sets as a special case of multisets (with multiplicity one for every element). As noted earlier, as long as we define types in terms of matrices, multisets and tuples, we can obtain the action on any compound type of other modelling languages in a similar way.

\begin{example}
Let $T$ be an unnamed type of size $3$, and let $X$ be of type \texttt{function (T $\rightarrow$ int(4..5))}. 
A possible value of $X$ is $x = \{ (1_T,4),(2_T,5),(3_T,4) \}$, representing the function that maps $1_T$ and $3_T$ to $4$, and $2_T$ to $5$.
Consider the permutation $g = \{ (1_T, 2_T)\}$ in $\Sym(T)$ which swaps $1_T$ and $2_T$ and leaves $3_T$ fixed.
Then the image of $x$ under $g$ is 
$x^g = \{ (1_T,4),(2_T,5),(3_T,4) \}^g = \{ (1_T,4)^g,(2_T,5)^g,(3_T,4)^g \}  = \{ (1_T^g,4^g),(2_T^g,5^g),(3_T^g,4^g) \} =   \{ (1_T^g,4),(2_T^g,5), \linebreak (3_T^g,4) \} = \{ (2_T,4),(1_T,5),(3_T,4) \} $, 
% \begin{align*}
%     x^g = &\{ (1_T,4),(2_T,5),(3_T,4) \}^g = \{ (1_T,4)^g,(2_T,5)^g,(3_T,4)^g \} \\
%     =& \{ (1_T^g,4^g),(2_T^g,5^g),(3_T^g,4^g) \} = \{ (1_T^g,4),(2_T^g,5),(3_T^g,4) \} 
%     \\ = &\{ (2_T,4),(1_T,5),(3_T,4) \},
% \end{align*}
representing a function that maps $2_T$ and $3_T$ to $4$, and $1_T$ to $5$, and where the subsequent expression rewriting uses Parts 4,5,2,1 of~\Cref{def: induced action} respectively.
\end{example}

One may find the presence of preimage in  matrix indexing in \Cref{def: induced action} to be unintuitive, but it is needed so that we have a group action, which is required for the symmetries to be well-behaved. One can also check that this definition is consistent to if we represent matrices as sets of tuples. 

\begin{example}
Consider a 1-dimensional matrix $m=[a,b,c]$ indexed by elements of unnamed types $[1_T, 2_T, 3_T]$. Let $g$ be the permutation $(1_T, 2_T, 3_T)$ and $h$ be $(1_T, 2_T)$. 
From \Cref{def: induced action}, we find that the image of $m$ first under $g$ and then under $h$ is $ ([a,b,c]^g)^h =  [c,a,b]^h = [a,c,b]$.
We get the same value if we take the image of $m$ under the composition of $g$ and $h$, as $gh = (2_T, 3_T)$ and $[a,b,c]^{gh} = [a,b,c]^{(2_T, 3_T)} = [a,c,b]$. 
However, if \Cref{def: induced action} had defined $(m^g)[i]$ to be $(m[i^g])^g$ instead, then $(m^g)^h = [b,c,a]^h = [c,b,a]$, which is not $m^{gh}$.
\end{example}

\subsubsection{Symmetries of Multiple Unnamed Types}

If a variable $X$ is constructed from multiple unnamed types, say $T$ and $U$, any combination of elements in $\Sym(T)$ and $\Sym(U)$ also permute $\Val(X)$. We say that the direct product $\Sym(T) \times \Sym(U)$ also acts on $\Val(X)$. In general, the direct product of groups  $G_1, G_2,  \ldots, G_k$ is the Cartesian product $G_1 \times G_2 \times \ldots \times G_k$ consisting of all $k$-tuple $(g_1, g_2,  \ldots, g_k)$ where each $g_i \in G_i$. 
If each $G_i$ acts on a set $\Omega_i$ and the $\Omega_i$'s are disjoint, the direct product $G_1 \times G_2 \times \ldots \times G_k$ acts on the disjoint union $\bigcup_{1 \leq i \leq k}\Omega_i$ by $\alpha^{(g_1, g_2, \ldots, g_k)} = \alpha^{g_i}$ if $\alpha \in \Omega_i$. %We may denote such a group by $\prod_{i=i}^k G_i$. 

\begin{definition} \label{def: action of DP}
Let $T_1, T_2, \ldots, T_m$ be distinct unnamed types.  
Then the direct product $D := \Sym(T_1) \times \Sym(T_2) \times \ldots \times \Sym(T_m)$ acts on $\Val(X)$, where $v^{(g_1, g_2, \ldots, g_m)} = ( \cdots ((v^{g_1})^{g_2})^{\ldots})^{g_m}$, for each element $(g_1, g_2, \ldots, g_m)$ of $D$ and $v \in \Val(X)$, and the application of each $g_i$ is as defined in \Cref{def: induced action}.   
\end{definition}

Recall that values of distinct unnamed types are disjoint sets. 
So, as the $g_i$'s permute distinct sets of points, they commute. That is, $g_ig_j=g_jg_i$ for all $i,j$.  
Then, since we have a group action, taking images under them is commutative since  
\( (v^{g_i})^{g_j} =  v^{(g_i g_j)} = v^{(g_j g_i)} = (v^{g_j})^{g_i} \)
for all $i,j$. 
So, the order in which we take the images when considering permutations of different unnamed types does not matter, as an extension to this, the order of unnamed types in the direct product also does not matter.

\begin{example}
Let $T$ and $U$ be unnamed types of size $2$ and $4$ respectively, and let $M$ be of type \texttt{matrix indexed by [$T$, int(1..3)] of $U$}. 
Then $D = \Sym(T) \times \Sym(U)$ acts on $\Val(M)$. 
Consider $m = [[1_U, 2_U,3_U],[2_U,3_U,4_U]] \in \Val(M)$ where we write the elements of $m$ in the order $T_1,T_2$, and $g=(1_T, 2_T)$ and $h = (1_U, 3_U)(2_U, 4_U)$. Here $g$ swaps $1_T$ and $2_T$, whereas $h$ and swaps $1_U$ and $3_U$ and also $2_U$ and $4_U$ at the same time. 
Then $(g,h) \in D$ and $ m^{(g,h)} = \left( [[1_U, 2_U,3_U], [2_U,3_U,4_U]]^{g} \right)^{h}$ by~\Cref{def: action of DP}, 
which gives $[[2_U,3_U,4_U]^{g}, $ \linebreak $  [1_U, 2_U,3_U]^{g}]^{h}   = [[2_U,3_U,4_U],[1_U, 2_U,3_U]]^{h}$ as $g$ permutes the indices and 
fixes values not from $T$. As $h$ fixes the indices $1_T,2_T$ and the indices $1,2,3$, this is just $[[2_U,3_U,4_U]^{h},  $ $ [1_U, 2_U,3_U]^{h}] = [[{(2_U)}^h,{(3_U)}^h,{(4_U)}^h], [{(1_U)}^h, {(2_U)}^h,{(3_U)}^h]] $. Finally, $h$ permutes values in $\Val(U)$, so  $m^g = [[4_U,1_U,2_U],[3_U, 4_U,1_U]]$.

% \begin{align*}
%     m^{(g,h)} &= \left( [[1_U, 2_U,3_U], [2_U,3_U,4_U]]^{g} \right)^{h} & \text{by~\Cref{def: action of DP}}   \\
%     &= [[2_U,3_U,4_U]^{g},[1_U, 2_U,3_U]^{g}]^{h} & \text{as $g$ permutes the indices} \\
%     &= [[2_U,3_U,4_U],[1_U, 2_U,3_U]]^{h} & \text{as $g$ fixes values not from $T$} \\
%     & = [[2_U,3_U,4_U]^{h},[1_U, 2_U,3_U]^{h}] & \text{as $h$ fixes the indices $1_T,2_T$} \\
%     &= [[{(2_U)}^h,{(3_U)}^h,{(4_U)}^h], &\\ 
%     & \qquad [{(1_U)}^h, {(2_U)}^h,{(3_U)}^h]] & \text{as $h$ fixes the indices $1,2,3$} \\
%     &= [[4_U,1_U,2_U],[3_U, 4_U,1_U]] & \text{as $h$ permutes $\Val(U)$}.
% \end{align*} 
\end{example}

% We say that a variable $X$ exhibits \emph{type symmetry} if there exist distinct values $x$ and $y$ in $\Val(X)$ such that $x^g = y$ for some $g \in G$. 
% In this case we say that $x$ and $y$ are \emph{symmetrically equivalent}. 
% As the name suggests, symmetrically equivalence forms an equivalence relation on $\Val(X)$.
% In particular, if $X$ is our decision variable, then we obtain an equivalence relation on the set of all solutions. 
% We say that the type symmetries are \emph{soundly} (resp. \emph{completely}) broken if at least one (resp. exactly one) solution per equivalence class is outputted.   
If $X$ is our only decision variable, then we obtain an equivalence relation on the set of all solutions, where two distinct solutions $x$ and $y$ in $\Val(X)$ are equivalent of $x^g = y$ for some $g \in G$. 
These symmetries are \emph{soundly} (resp. \emph{completely}) broken if at least one (resp. exactly one) solution per equivalence class is outputted.  
If we have multiple decision variables $V_1, V_2, \ldots, V_d$, then we can reduce to the case where there is only one decision variable, which is the tuple $(V_1, V_2, \ldots, V_d)$. 
This is particularly important when we want to ensure the consistent application of permutations of unnamed types across multiple variables, such as in the template design problem we have seen in \Cref{sec: example models}.

\subsection{Breaking the Symmetries of Unnamed Types}
\label{sec: type symmetries}
\label{section: break typ sym}

A common and general way to break symmetries is to use the lex-leader constraints~\cite{crawford1996symmetry}.
In general, for a group $G$ acting on the domain $\Val(X)$ of a variable $X$, the (value) lex-leader constraint $LL_{\symleq}(G,X)$ for $X$ under $G$ with respect to a total ordering~$\symleq$ of $\Val(X)$ is the constraint $\forall \sigma \in G. \, X \symleq X^\sigma$. 
This asserts that, in any $G$-induced equivalence class of $\Val(X)$, only one value (the smallest one) can be assigned to $X$. 
Recall that we treat our problems as containing a single variable, which may be a tuple, so we need only consider symmetry breaking for a single variable $X$.
%A constraint is a \emph{sound} symmetry breaking constraint for $X$ under a symmetry group $G$ it allows at least one symmetric image of each assignment to $X$. A constraint  is a \emph{complete} symmetry breaking constraint is it allows at most one symmetric image of each assignment to $X$.
In this paper, we always base our symmetry breaking on a lex-leader constraint $LL_{\symleq}(G,X)$. In this situation, sound symmetry breaking constraints will be implied by $LL_{\symleq}(G,X)$, and complete symmetry breaking constraints will imply $LL_{\symleq}(G,X)$ as well.

%A constraint is a \emph{sound} symmetry breaking constraint for variable $X$ under a symmetry group $G$ if it is implied by $LL_{\symleq}(G,X)$ for some total ordering $\symleq$ on $\Val(X)$. 
%Such a constraint is said to be \emph{complete} if it also implies $LL_{\symleq}(G,X)$.  

To completely break the unnamed type symmetries of a variable $X$, we use $LL_{\symleq}(G, X)$, where $G$ is the unnamed type symmetry group acting on $\Val(X)$ and $\symleq$ is a total ordering on $\Val(X)$. 
So we can eliminate unnamed type symmetries in the following way, the proof of which follows from the correctness of lex-leader constraints eliminating all but one solution in each equivalence class.

\begin{proposition} \label{prop: can get model withou symmetries using lex leaders}
Starting with an \essence model $M$ with unnamed types \linebreak $T_1, T_2, \ldots, T_k$ of size $s^1, \ldots, s^k$ respectively, we can obtain an equivalent model (in the sense that there is a bijection between the solution sets) \emph{without} unnamed types in the following way. Letting $V_1,V_2, \ldots, V_d$ be the decision variables of $M$, first replace each unnamed type $T_i$ by an enumerated type with values $1_{T_i}, 2_{T_i}, \ldots, s^i_{T_i} $. Then let $X$ be a new decision variable representing the tuple $(V_1, V_2,  \ldots , V_s)$. Finally, we add the lex-leader constraint $LL_{\symleq}(\Sym(T_1) \times \Sym(T_2) \times \ldots \times \Sym(T_k), X)$, where $\symleq$ is a total ordering on $\Val(X)$.    
\end{proposition}

Therefore, we need a total ordering of $\Val(T)$ for any type $T$ that is not constructed from unnamed types. %Then, since the set $\Val(X)$ of all possible values of a variable $X$ of type $T$ is a subset of $\Val(T)$, we also obtain a total ordering on $\Val(X)$. 
We shall define these orderings in the next subsection. 
This will then inform us on how we can refine constraints of the form $X \symleq Y$ when $X$ and $Y$ are of abstract types.

Before we move on, note that there are many papers which consider different methods of generating subsets of symmetry breaking constraints (see \cite{GENT2006329} for an overview). 
The most common technique is to replace $LL_{\symleq}(G, X)$ with $LL_{\symleq}(S, X)$ for some subset $S$ of $G$, to obtain sound but incomplete symmetry breaking constraints.  
When $G$ is a direct product $G_1 \times G_2$, we may use $LL_{\symleq}(G_1 \cup G_2, X)$ instead.
When $G$ is the symmetric group $\Sym(\{x_1, x_2,\ldots, x_n\})$, we can take $LL_{\symleq}(S, X)$ where $S = \{ (x_i,x_j) \mid 1 \leq i,j \leq n\}$ consists of all permutations in $G$ that swap two elements. Alternatively, we can take $S$ to be $ \{ (x_i,x_{i+1}) \mid 1 \leq i \leq n-1 \}$, the set of all permutations in $G$ that swaps any consecutive points. 
%These are the core ideas behind ``double-lex'' \cite{rowColSym}, which deals with row and column symmetries of matrices by ordering the rows and the columns. There have been a number of extensions to this idea, including breaking adding other sets of constraints \cite{10.1007/978-3-540-45193-8_22}

%\subsection{Ordering of compound and abstract types}
\subsubsection{Total Ordering for All Types}
\label{section: fixed ordering}

Here, we describe a general approach to defining a total ordering $\symleq_T$ of $\Val(T)$ for any given type $T$ not constructed from unnamed types. To simplify notations, we drop the subscript $T$ where doing so will not cause confusion.
The actual ordering used does not matter for correctness, as long as it is a total order.  
We first define an order on multisets in terms of the ordering on its members' type.  We then show how ordering on other types can be defined in terms of the multiset ordering.  This is the ordering used in our implementation in \conjure. 
%For a total ordering $\leq$ on set $A$, the \emph{lexicographical ordering (or lex-ordering) $\leq_{lex}$ over $\leq$} is the total ordering on the tuple-like structures over $A$ such that $(t_1,t_2, \ldots, t_k) \leq_{lex} (t'_1, t'_2, \ldots, t'_k)$ if and only if either $t_i = t'_i$ for all $i$, or there is an $i$ such that $t_i \leq t'_i$ and $t_j = t'_j$ for all $j < i$.

For the ordering on multiset, we used an ordering very similar to one in the literature~\cite{multisetOrdering,kiziltanMultisets}. We assume that the type $M$ is a multiset of elements of type $T$ and that the ordering $\symleq_T$ is defined. 
We say that $m_1 \symleq_M m_2$ if and only if: $m_2 = \emptyset$, or; $m_1, m_2 \neq \emptyset$ and $\min(m_1) \symless_T \min(m_2)$, or; $m_1, m_2 \neq \emptyset$ and $\min(m_1) = \min(m_2)$ and $m_1 \setminus \{ \min(m_1) \} \symleq_M m_2 \setminus \{ \min(m_2) \}$.
This ordering may be unintuitive, but it is chosen so that ordering on multisets is equivalent to lex-ordering of a natural representation (specifically the occurrence representation; see, for example, \cite{kiziltanMultisets} for proof). 
Since multisets are abstract types, constraints $X \symleq_M Y$, when $X$ and $Y$ are multisets, will need to be refined, and we can do so using this occurrence representation. 
The ordering for all other types can be found in the following definition.

\begin{definition}\label{def: ordering compound types}
Let $T$ be an \essence type not constructed from any unnamed types. We define a total ordering $\symleq_T$ for values $\Val(T)$ of type $T$ recursively by:
\begin{enumerate} \label{def: ordering on semantics of types}
    \item if $\Val(T)$ consists of integers, we take $\symleq_T$ to be $\leq$ on integers;
    \item if $\Val(T)$ consists of Boolean, we use $\texttt{false} \symleq_T \texttt{true}$;
    \item if $\Val(T)$ consists of enumerated types, then $x \symleq_T y$ if $x$ occurs before $y$ in the definition of the enumerated type;
    \item if $\Val(T)$ consists of matrices or tuples of an inner type $S$, then take $\symleq_T$ to be lexicographical order $\symleq_{{lex}}$ over an order $\symleq_S$ for the inner type; 
    \item if $\Val(T)$ consists of multisets, take $\symleq_T$ to be the multiset ordering $\symleq_{M}$ above.
\end{enumerate}
%The ordering for all other types are obtained using \Cref{remark: values of types as sets and tuples}. 
\end{definition} 

Note again that using \Cref{def: semantics of types}, this definition gives an ordering for \emph{all} \essence types. 
An ordering for all types in other modelling languages can be defined in a similar way, by defining compound types in terms of multisets, tuples and matrices, and defining a concrete total ordering on each atomic type.

\begin{remark} \label{remark: rewriting using static ordering}
We can therefore refine constraints of the form $X \symleq Y$ to: $X \leq Y$ for the appropriate atomic ordering $\leq$ if $X$ and $Y$ are atomic; $X \symleq_{lex} Y$ if $\Val(X)$ and $\Val(Y)$ are matrices or tuples, and to 
$[-\mathrm{freq}(X,i) | i \in X] \symleq_{lex} [-\mathrm{freq}(Y,i) | i \in X]$, using a ordering of $X$, if they are (multi)sets, where $\mathrm{freq}(X,i)$ gives the number of occurrence of $i$ in~$X$.  
As the base cases in \Cref{def: ordering compound types} are ordering $\leq$ of atomic types, these $\symleq_{lex}$ will eventually be rewritten to $\leq_{lex}$. 
\end{remark}

\section{Implementation in Conjure}
\label{sec: implementation}

%Having seen what are unnamed type symmetries and how we can theoretically break them, we now see how these are implemented in \conjure and present some case studies. 

%\subsection{Implementation summary}

%\todo{say,  due to space, will have more implementation details in subsequent paper.}
% In this section we give a  summary of how the symmetry breaking method described in this paper is implemented in \conjure. %A detailed account of the implementation is out of the scope of this paper. 

We outline the implementation of the symmetry breaking method described in this paper within \conjure. The complete implementation can be found in the \conjure repository at \url{https://github.com/conjure-cp/conjure}.

\subsubsection{Permutations}
We introduce a new \essence type \texttt{permutation}.
We allow permutations of integers, enumerated types or unnamed types. 
Permutation is available as a domain constructor, using keywords \texttt{permutation of}, and takes either an integer, enumerated or unnamed type domain as argument. 
Permutation values and expressions are written in cycle notations. 
Each permutation has an attribute \texttt{NrMovedPoints}, which gives the number of points that are not fixed by the permutation.

Permutations can be naturally represented as bijective total functions, which in turn can be represented as 1-dimensional matrices, where the element at a certain index gives the image of the index.  
%As permutations are bijective (total) functions, a (known) permutation $g:= (a_{11}, \ldots, a_{1k_1})  \ldots (a_{r1}, \ldots, a_{1r_1})$ are represented with a matrix $M$ of $X:=\bigcup_{i,j} a_{ij}$ indexed by $X$, where $M[a_{ir_i}]=a_{i1}$ and $M[a_{ij}]=a_{i(j+1)}$ for all other $j$. 
Each permutation is stored with its inverse. This is because it turns out we almost always need to use the inverse of a permutation during symmetry breaking, and storing both the permutation and its inverse was the most efficient option in practice. The operator $\texttt{permInverse}$ gets the inverse of a permutation. $\texttt{permInverse}$ uses the fact that the inverse of the inverse of a permutation is the original permutation, so there is no need to calculate any further permutations when calling \texttt{permInverse}.
%The inverse is represented by a matrix $M'$ of $X$ indexed by $X$, where $M[a_{i1}]=a_{ir_i}$ and $M[a_{ij}]=a_{i(j-1)}$ for all other $j$. 

There is an operator $\texttt{image}$ that takes a permutation $g$ on a type $T$ and a value $x$ of type $T$ such that $\texttt{image}(g,x)$ gives the image of $x$ under $g$. 
The more general operator $\texttt{transform}(g, X)$ represents the image of the induced action of $g$ on the values of its second argument. If $g$ is a permutation on a type $T$, and the type of $X$ contains no reference to $T$, then $\texttt{transform}(g, X) = X$. 
%unknown case: (e.g. when perm is a var) we represent using a set of tuples, where tuple (f,t) indicates that the permutation maps f to t. 

\subsubsection{Unnamed Types}
Unnamed types domains, similar to enumerated types, are eventually converted to $\texttt{int(1..s)}$ where $s$ is the size of the unnamed type. During refinements, since we need to keep track of the original type of these new integers (e.g. for permutation applications), unnamed types are converted to \emph{tagged integers}, which behaves like integers but remembers the name (\texttt{IntTag}) of the unnamed type it comes from. 
Therefore, we start by removing the declarations of the unnamed types (\texttt{letting $T$ be new type of size $s$}). For all other domains constructed using such a type $T$, we replace them with tagged integers $1, \ldots, s$, labelled with~$T$. 

Permutations on an unnamed type $T$ are refined into permutations on integers tagged with $T$, and \texttt{image} and \texttt{transform} will only change integer values or variables with the appropriate tag. The refinement rules of \conjure ensure that when tagged integers are refined, the tag is preserved. For example a set of a tagged integers of type $T$ is refined into a matrix indexed by $T$ of booleans.

%The TypeInt constructor has been extended to carry a TagInt type that can be constructed by \texttt{TagInt | TagUnnamed Text | TagEnum Text | TaggedInt Text}. Regular integers will be of type \texttt{(TypeInt TagInt)}, Enums will be converted to type \texttt{(TypeInt (TagEnum "enumname"))}, Unnameds will be converted to type \texttt{(TypeInt (TagUnnamed "unnamedname"))}, and integers with a supplied tag will have type \texttt{(TypeInt (IntTag "tagtext"))}. The type checker now takes an implicit argument enabling ad-hoc switching between Strong and RelaxedIntegerTags modes. When we type check strongly we expect all aspects of the integer type to match. This means that integer operations will not typecheck for Enum or Unnamed integers in strong mode. When we relax type checking we ignore the TagInt value and allow integer operations to work across integer variants. Relaxed typechecking therefore allows us to provide optimising rewrite rules that can treat Unnameds and Enums as their underlying Integer representation. Permutation image rewrite rules use Strong type unification on integers such that a permutation on an unnamed type will only permute integers with that integer tag.

\subsubsection{Symmetry Breaking Constraints}

As discussed in \Cref{section: break typ sym}, we break the symmetries induced by unnamed types using lex-leader constraints. 
Our implementation allows sound but incomplete symmetry breaking by replacing the group in $LL$ with a subset. 
%Exactly which subsets to take is determined by flags. 
The actual lex-leader constraints added depend on the run-time flags. 
%To vary the permutations used in the lex-leader contraints, we use  \texttt{Consecutive}, \texttt{AllPairs}, or \texttt{AllPermutations} flags. 
%The \texttt{Independently} and \texttt{Altogether} flags vary how we deal with multiple unnamed types. 
%In all instances, we add lex-leader constraints on the tuple of all decision variables. 
For a model with unnamed types $T_1,T_2, \ldots, T_k$ and decision variables $V_1, V_2, \ldots, V_s$, the  constraint used is $\bigwedge_{1 \leq i \leq k} LL(S_i, (V_1,  V_2, \ldots, V_s) )$ if  with the \texttt{Independently} flag; and if $LL(S_i \times \ldots \times S_k, (V_1, V_2, \ldots, V_s) )$ if with the \texttt{Altogether} flag, 
where each $S_i$ is a subset of $\Sym(T_i)$ such that   $S_i := \{ ( j_{T_i} , (j+1)_{T_i} ) \mid 1 \leq j < |T_i|\}$ if with the \texttt{Consecutive} flag; $S_i := \{ (t,u) \mid t,u \in T_i\}$ with the \texttt{AllPairs} flag; and  $S_i := \Sym(T_i)$ with the \texttt{AllPermutations} flag.
Each $LL(S,X)$ is expressed as the conjunction, over all possible permutations $g \in S$, of expressions of form $X \, \texttt{.<=}  \, \texttt{transform}(g, X)$.
Here $\texttt{.<=}$ represents the total order $\symleq_T$ from \Cref{def: ordering compound types}, where $T$ can be any suitable type.  

% Each $LL(S,x)$ is expressed as the conjunction, over all possible permutation $g \in S$, of expressions of form depending of a further flag: 
% \begin{itemize}
%     \item \texttt{Quick} flag: $\texttt{quickPermOrder}(g,x)$, where $\texttt{quickPermOrder}$ signifies that further refinements are to be done using the method described in \Cref{sec: fast sym break};
%     \item \texttt{Complete} flag: $x \, \texttt{.<=}  \, \texttt{transform}(g, x)$,  where $\texttt{transform}(g, x)$ represents the image of the induced action of $g$ on values of its second argument. 
% \end{itemize}
% We also have the \texttt{fast} flag as an alias for \texttt{Quick-Consecutive-Independently} and the \texttt{full} flag for \texttt{Complete-AllPermutations-Altogether}. 

\subsubsection{Permutation Application}

%In the \texttt{Complete} case, w
%From above, we have expressions of the form \linebreak  $X \, \texttt{.<=} \,\texttt{transform}(g, X)$, where $g$ is a permutation and $X$ an expression. 
Starting from  $X \, \texttt{.<=} \,\texttt{transform}(g, X)$ from above, if $g$ is a list of permutations $[g_1, g_2, \ldots, g_r]$ representing an element of a direct product, we rewrite expressions of the form  $X \,\texttt{<=} \, \texttt{transform}([g_1, g_2, \ldots, g_r], X)$ to the conjunction of 
$X \, \texttt{.<=}\,  x_r$ and $x_i = \texttt{transform}(g_i, x_{i-1})$ for $1 \leq i \leq k$, where $x_0$ is simply $X$ and the $x_i$'s are new variables. 
The refinement rules for $\texttt{transform}(g,x)$, when $g$ is a permutation, follow directly from the definition of permutation application from \Cref{def: induced action}. \conjure's general design, which applies rewrite rules until all high-level types and operators are removed, can easily handle this new set of rules.
As an example, for a matrix $X$, we rewrite each entry $\texttt{transform}(g, X)[i]$ to $\texttt{transform}(g, X[\texttt{transform}(\texttt{permInverse}(g), i)])$. These internal \texttt{transform} and \texttt{permInverse} are further refined, until all permutations have been removed, and only types in \essencePrime remain.

\subsubsection{Refining Ordering Constraints}

Each constraint of the form $X \texttt{.<= } Y$ is refined to $ \texttt{symOrder}(X)  \texttt{.<= } \texttt{symOrder}(Y)$. Here $\texttt{symOrder}(X)$ signifies that we are to rewrite the expression using \Cref{remark: rewriting using static ordering}. 
So $X \texttt{.<= } Y$ will eventually be written to expressions of the form $ X' \texttt{ <=lex } Y'$ or $X' \texttt{ <= } Y'$ for some $X'$ and $ Y'$, where \texttt{<=} is the order of atomic types and \texttt{<=lex} the lexicographic ordering over \texttt{<=}. 
%The implementation of this in \conjure predates this project on unnamed type. 
The lexicographic constraints will typically contain every variable, but can often be simplified, e.g.\ $[a,b,c,d] \leq_{lex} [a,d,b,d]$ can be simplified to $[b,c] \leq_{lex} [d,b]$. Rather than perform these simplifications while initially generating and refining the lexicographic ordering constraints, a set of general rules for simplifying lexicographic ordering constraints \cite{ohrman2005} is run after refinement is finished.

\begin{table}[t]
\centering
\caption{How the unnamed types occur in some problems}
\label{table: unnamed type uses in csplib problems}

\begin{tabular}{|l|l|}
\hline
Problem&Type\\
\hhline{==}
Lam's Problem~\cite{csplib:prob025}& \texttt{matrix indexed by [$T$,$T$] of ?} \\ 
\hline
Set-theoretic Yang-Baxter~\cite{akgun2022YB} &  \texttt{matrix indexed by [$T$,$T$] of $T$} \\
\hline
%Balanced Incomplete Block Design~\cite{csplib:prob028}& relation of $T_1$ * $T_2$\\
Balanced Incomplete Block Design~\cite{csplib:prob028}& \texttt{matrix indexed by  [$T_1, T_2$] of ?}\\
\hline
%Social Golfers~\cite{csplib:prob010}& set of partition from $T$\\
Social Golfers~\cite{csplib:prob010}& \texttt{matrix indexed by [$T_1, T_2$] of $T_3$} \\
\hline
%Covering Array~\cite{csplib:prob045}& mset of function (total) $T$ $\rightarrow$ $T$ \\
Covering Array~\cite{csplib:prob045}& \texttt{matrix indexed by [$T_1, T_2$] of $T_3$} \\
\hline
%Template Design~\cite{csplib:prob002}& \makecell[l]{ function (total): $(T,?)$ $\rightarrow$ ? \\  function (total): $T$ $\rightarrow$ ?} \\
Template Design~\cite{csplib:prob002}& \makecell[l]{ \texttt{matrix indexed by $[T_1]$ of ?} \\  \texttt{matrix indexed by $[T_1, T_2]$ of ?} } \\
\hline
Rack Configuration~\cite{csplib:prob031}& \texttt{function $T$ $\rightarrow$ ?}\\
\hline
Semigroups~\cite{semigroup}& \texttt{function $(T,T)$ $\rightarrow$ T}\\
\hline
Vellino's Problem~\cite{csplib:prob116}& \makecell[l]{ \texttt{function: $T_1$ $\rightarrow$ $T_2$} \\ \texttt{function: $T_1$ $\rightarrow$ mset of $T_3$}}\\
\hline
Sports Tournament Scheduling~\cite{csplib:prob026}&  \texttt{relation of ($T_1$ * $T_2$ * set of $T_3$)}\\
\hline
%Production Line Sequencing~\cite{csplib:prob131}& \makecell[l]{ function (total) $T$ $\rightarrow$ ?\\function (total) $T$ $\rightarrow$ ? \\function (total) ? $\rightarrow$ $T$}\\
%\hline
\end{tabular}
\end{table}

%Now we can go through the refinement process as usual. As we have expressions of the form $X \texttt{.<=} Y$, we can use the same representations for both $X$ and $Y$. So these expressions will be rewritten using the dynamic ordering from \Cref{section: dynamic ordering}, so will eventually rewritten to lex ordering, or ordering on atomic types. %\todo{say we add new global variables and equality constraint?}

% In the \texttt{Quick} case, expressions of form  $\texttt{quickPermOrder}((g_1, \ldots, g_r),x)$ are rewritten to 
% % the conjunction of 
% % \linebreak $\texttt{symOrder}(X) \, \texttt{.<=}\,  \texttt{symOrder}(x_r)$ and $x_i = \texttt{transform}(g_i, x_{i-1})$ for $1 \leq i \leq k$ where $x_0$ is now $\texttt{symOrder}(X)$. 
% \begin{align*}
%    & \texttt{symOrder}(x)  \texttt{.<=} \\
%    & \qquad \texttt{transform}(g_r, (\texttt{transform}(g_{r-1}, ( \ldots   (\texttt{transform}(g_1, \texttt{symOrder}(x)) \ldots)) )).
% \end{align*}
% Here $\texttt{symOrder}(x)$ signifies that we are to use fixed static ordering in \Cref{def: ordering compound types}. So we will rewrite the above constraint using \Cref{remark: rewriting using static ordering}. 

\section{Case Studies and Discussion}
\label{sec: results}
\label{sec: discussion}

We give a few case studies to demonstrate our implementation of the method. 
%In particular, we will consider several cases where the decision variables are all of matrix type, but with varying number of unnamed types occurring in various positions.
Future work will include more in-depth experimentation. 
We consider the problems from \Cref{sec: example models}, and three further problems with matrix types as decision variables, but with varying number of unnamed types occurring in various positions, as well as some examples where the decision variables are of different types. 
The types of the decision variables of these problems are summarised in \Cref{table: unnamed type uses in csplib problems}, where the $T$ and $T_i$'s are all distinct unnamed types, and `?' denotes other types that are not constructed from any unnamed types.
%type \texttt{matrix indexed by  [$T_1, T_2$] of ?}, \texttt{matrix indexed by [$T$,$T$] of ?} and \texttt{matrix \linebreak indexed by [$T_1, T_2$] of $T_3$} respectively, where the $T$ and $T_i$'s are all (distinct) unnamed types, and `?' denotes other types that are not constructed from any unnamed types.
The models in \essence, and the automatically generated \essencePrime models, for all combinations of flags, can be found at \url{https://github.com/stacs-cp/CPAIOR2025-Symmetry}. 
The resulting models were manually inspected for correctness. 
%, and are used to solve some small instances and the (number of) solutions are verified for correctness. 
In particular, the number of solutions for small instances of the set-theoretic Yang-Baxter equation and the semigroup problem are consistent with those in the literature~\cite{oeisYB,oeisSemigroup}. 

The different methods of symmetry breaking provide an easy way of choosing between different tradeoffs. \texttt{Altogether-AllPermutations} will break all symmetries, producing an exact list of symmetry-broken solutions, at the cost of a very large number of constraints. The fact that we need many constraints is not surprising, as the symmetries of unnamed types include several cases which have been proved theoretically difficult.
Consider the solutions to a problem with a decision variable with type \texttt{matrix indexed by [T,T] of bool}, with no constraints. These solutions can be viewed as directed graphs, so completely breaking the unnamed type symmetries is equivalent to finding the canonical image of these graphs. Similarly, a \texttt{set of set (size 2) of T} can be viewed as the edges of an undirected graph. Checking if two solutions of these problems are equivalent is therefore as hard as the graph isomorphism problem. 
Further, a variable of type \texttt{matrix indexed by [T1,T2] of bool} has row and column symmetries, and efficient generation of complete symmetry breaking constraints is also at least as hard as the graph isomorphismm problem~\cite{anders2024complexitysymmetrybreakinglexleader}.

If we consider a matrix indexed by two unnamed types, e.g. in the balanced incomplete block design problem, then \texttt{Independently-Consecutive} produces ``double-lex'', one of the most widely used symmetry breaking methods~\cite{rowColSym}. The constraints generated by \texttt{Independently-AllPairs} have also been used in practice \cite{10.5555/647489.727148}, and can lead to faster solving. Deciding exactly what level of symmetry breaking is best for a particular class of problem is future work.

% These row and column symmetries can be efficiently broken using ``double-lex'' \cite{rowColSym}, which deals with these symmetries by ordering the rows and the columns.
% While is is not difficult to hard case \conjure to handle these cases by giving the double-lex constraints, it is an unsatisfactory solution. 
% We are therefore investigating how we can automatically generate incomplete but more efficient symmetry breaking constraints(see next section for details), which includes the double-lex, or their extensions~\cite{10.1007/978-3-540-45193-8_22}, in the case where there are row and column symmetries. 
% Nevertheless, the result of this paper serves as a ``gold standard" to compare our future work against. 

% For this this problems with this this setting, we get model consistent with literature/CSPlib?

% complete vs fast 

% fixed vs dynamic ?

%see: https://github.com/stacs-cp/CP2019-Symmetries
%\todo{maybe Do say earlier that unnamed symemtries are a mystery - we don't know if it's var /val/soln/... symemtries, and in fact they can be all. We didn't even know how to define it! But  }
%\todo{explicitly say which cases are row and colm syms, which is GI (cite?). Say can try to detect and hard case it, but we are investigating other generalisable methods...}

\section{Conclusion and Future Work}

In this paper, we show how the symmetries of indistinguishable objects can be broken completely together with an implementation in \essence. We do so by introducing the new \texttt{permutation} type, as well as a total ordering for all possible types in \essence, which allows us to express and automatically generate symmetry breaking constraints for unnamed types.
We have also seen how we can soundly but incompletely break unnamed type symmetries by controlling the permutations used in the lex-leader constraints. 
Our abstract treatments of types (using multisets, tuples and matrices) and unnamed type symmetries (in terms of group actions) make our method generalisable to any other solving paradigms with indistinguishable objects. This paper also serves as a theoretical background for further work in this area. 
%\todo{want to say that all future papers will depend on this theory (which is already very long), and will also be generalisable to other paradigms, sort of to motivate this primarilt theoretical paper. }

Much further work awaits. The symmetry breaking method here can be prohibitively expensive in some cases. 
%in particular when we have 2-dimensional matrices with row and column symmetries. 
We therefore will investigate how some relaxations of the ordering constraints can be used to give faster symmetry breaking method. 
%In particular, we aim to automatically generate the efficient `double-lex' method for the case of 2D matrices with row and column symmetries \cite{rowColSym}, and generalisation of that for higher-dimensional matrices.
Furthermore, the static ordering described in \Cref{section: fixed ordering} can also be a source of inefficiency. 
This is because, when rewriting multisets to their occurrence representations in \Cref{remark: rewriting using static ordering}, the elements in $[-\mathrm{freq}(X,i) | i \in X]$ must be sorted according to the total ordering of the types of the inner set elements. This can be particularly difficult when we have deeply nested types. 
In general, it is not possible to produce a single global ordering which can be refined to a simple and efficient set of constraints in all possible representations.
We shall investigate the use of representation-specific total orderings. 
%We shall investigate how a representation-specific method of creating a total ordering can give an alternative total ordering that circumvents this problem.. 

Note that the symmetry breaking techniques in this paper will work even when the symmetry groups on unnamed types are not necessarily symmetric groups. 
We shall explore how the new permutation type can be used by an expert user for more control on symmetry breaking, in allowing the action of arbitrary permutation groups. 
It is worth exploring how to automatically generate efficient symmetry breaking constraints for commonly occurring permutation groups, such as the chessboard symmetry. We would also like to explore symmetry breaking for other group actions, e.g. actions on non-atomic types. 
%In terms of solving, we will look at breaking these symmetries dynamically, e.g. by influencing branching choices. 

Generally, we intend to perform a more extensive analysis of the symmetry breaking constraints produced from our method (and future extensions), comparing it to the custom ones from the literature, and exploring other ways of breaking symmetries that are consistent with the best known methods in the literature.

%We are also interested in seeing how the new permutation type can be used by an expert user for more control on symmetry breaking, and in allowing the action of arbitrary permutation groups. Note that the symmetry breaking techniques in this paper will work even when the symmetry group on unnamed types are not the symmetric groups. 

\bibliographystyle{splncs04}
\bibliography{bib}

\newpage

\end{document}